\documentclass[letterpaper]{article} 
\usepackage{aaai2027}  
\usepackage[hyphens]{url}  
\usepackage{graphicx} 
\usepackage{natbib}  
\usepackage{caption} 
\usepackage{algorithm}
\usepackage{algorithmic}

\usepackage{newfloat}
\usepackage{listings}
\DeclareCaptionStyle{ruled}{labelfont=normalfont,labelsep=colon,strut=off} 
\floatstyle{ruled}
\newfloat{listing}{tb}{lst}{}
\floatname{listing}{Listing}

\usepackage{booktabs}

\usepackage{amsmath}
\usepackage{multirow}
\usepackage{amssymb}
\usepackage{lipsum}

\title{QR-Erase: Efficient Subspace-Based Machine Unlearning with Layer Localization}
\author {
    Tyler Lizzo,
    Larry Heck,
}
\affiliations {
    Georgia Institute of Technology\\
    lizzo@gatech.edu, larryheck@gatech.edu
}

\begin{document}

\maketitle

\begin{abstract}
Machine unlearning seeks to remove targeted information from trained models without requiring costly retraining. Existing optimization-based methods often degrade unrelated capabilities, while subspace-based approaches rely on computationally expensive singular value decompositions (SVD). We introduce \textsc{QR-Erase}, a subspace-based framework that uses Pivoted QR decomposition to identify and remove task-specific representations directly from model parameters. We further propose Layer-Localized \textsc{QR-Erase}, which restricts updates to layers containing the highest concentration of task-specific information. We show that Pivoted QR provides accurate subspace recovery with bounded error, and that under a mild spectral gap condition, the recovered subspace approaches the optimal SVD solution. Across task-level, cross-lingual, and speech unlearning, \textsc{QR-Erase} achieves a stronger forgetting--retention tradeoff than optimization-based methods while remaining within \textbf{5\%} of SVD across all metrics. Exploiting low-rank and layer-localized structure further improves forgetting (for example, reducing speech forget-set accuracy from 53.1\% to 15.7\%). These results demonstrate that accurate subspace recovery, rather than optimal reconstruction, is sufficient for effective unlearning and provides an efficient and general alternative to SVD-based methods for modern foundation models.
\end{abstract}


\section{Introduction}
Machine unlearning seeks to remove the influence of specific data, concepts, or capabilities from a trained model while preserving performance on unrelated tasks~\cite{bourtoule2021machineunlearning,lizzo2026unlearningllmsmethodsevaluation}. Growing concerns surrounding privacy, data ownership, and regulations such as GDPR~\cite{gdpr2016} and CCPA~\cite{ccpa2019} have made efficient machine unlearning increasingly important for modern foundation models.

Most existing approaches rely on optimization-based updates, including gradient ascent, retain-objective optimization, and preference-based learning. While effective for forgetting, these methods often degrade unrelated capabilities and require iterative optimization that scales poorly to large models. More recently, subspace-based approaches have represented task-specific knowledge within low-dimensional parameter subspaces, enabling direct removal in weight space. Although promising, these methods raise two fundamental questions: \emph{Is optimal low-rank reconstruction necessary for effective machine unlearning?} and \emph{Is task-specific knowledge uniformly distributed throughout a model, or concentrated within specific layers?}

In this work, we investigate both questions through two complementary design choices. First, we introduce \textsc{QR-Erase}, which replaces singular value decomposition (SVD) with computationally efficient Pivoted QR for task subspace recovery. This allows us to evaluate whether accurate subspace recovery, rather than optimal low-rank reconstruction, is sufficient for effective machine unlearning. Second, we propose a layer-localized forgetting strategy that restricts updates to layers exhibiting above-average task energy, enabling us to study whether identifying \emph{where} knowledge is represented is more important than applying forgetting uniformly across the network. Together, these complementary contributions disentangle the roles of efficient subspace recovery and knowledge localization in subspace-based machine unlearning.

We evaluate these ideas on task-level and factual text unlearning, cross-lingual factual unlearning, and speech unlearning through sample- and speaker-level forgetting. These settings span language, multilingual, and speech representations, allowing us to evaluate whether the same principles of subspace recovery and layer localization generalize across diverse forms of learned knowledge.

\begin{figure*}
    \centering
    \includegraphics[width=.9\linewidth]{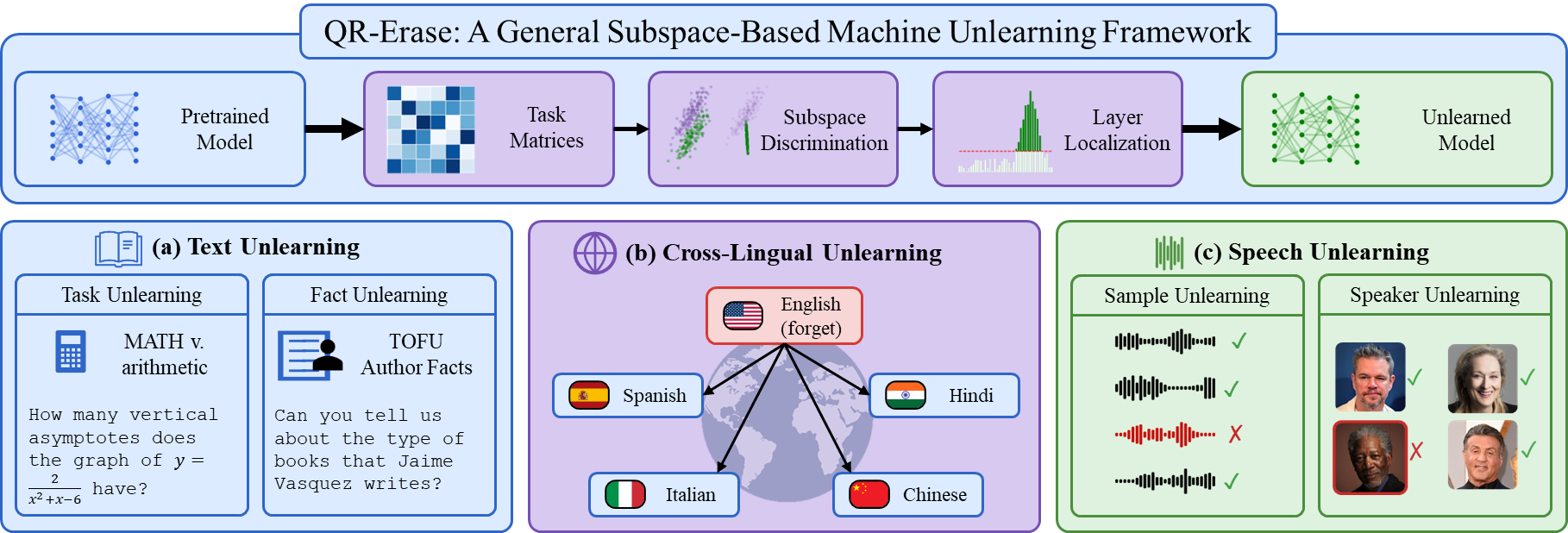}
    \caption{ \textbf{Overview of \textsc{QR-Erase} and evaluation settings.} We introduce \textsc{QR-Erase}, a Pivoted QR-based machine unlearning framework with layer-localized removal, and evaluate it across three complementary settings: (a) task-level and factual text unlearning, (b) cross-lingual factual unlearning, and (c) speech unlearning through sample and class forgetting.}
    \label{fig:overview}
\end{figure*}

\begin{itemize}
    \item We investigate whether optimal low-rank reconstruction is necessary for effective machine unlearning by introducing \textsc{QR-Erase}, a Pivoted QR-based alternative to SVD that achieves comparable forgetting performance while reducing decomposition runtime by over 60\%.

    \item We introduce a layer-localized forgetting strategy based on normalized task energy and show that restricting updates to high-energy layers consistently improves the forgetting--retention tradeoff across subspace-based unlearning methods.

    \item We evaluate \textsc{QR-Erase} and layer-localized forgetting across text, multilingual, and speech machine unlearning, demonstrating that the two design choices provide complementary benefits: Pivoted QR improves computational efficiency, while layer localization drives improvements in unlearning performance.

    \item We provide new empirical evidence that effective machine unlearning depends on accurately recovering task-specific subspaces and identifying where knowledge is localized within modern foundation models.
\end{itemize}

\section{Related Work}
\label{sec:related_work}

\subsection{Machine Unlearning}

Machine unlearning aims to remove the influence of specific data, concepts, or capabilities from a trained model while preserving performance on unrelated tasks~\cite{lizzo2026unlearningllmsmethodsevaluation}. Exact approaches retrain models without the targeted data but are impractical for modern foundation models due to their computational cost~\cite{bourtoule2021machineunlearning}. Consequently, most existing methods perform approximate post hoc updates to model parameters or training objectives. Optimization-based approaches include Gradient Ascent~\cite{jang2023knowledge}, Gradient Difference~\cite{liu2022continual}, SCRUB~\cite{scrub2023}, Bad Teaching~\cite{badteaching2023}, KL-based retain objectives~\cite{wang2025balancing}, and preference-based methods such as DPO and NPO~\cite{rafailov2024directpreferenceoptimizationlanguage,zhang2024npo}. While effective, these methods often exhibit a tradeoff between forgetting and retained capability.

\subsection{Subspace-Based Unlearning}

Parameter-space approaches manipulate directions associated with learned behavior. Task arithmetic showed that task-specific capabilities can be represented as vectors in weight space~\cite{ilharco2023editing}, while model-editing methods such as ROME and MEMIT perform localized updates to factual associations~\cite{meng2022locating,meng2023memit}. More recently, projection-based and geometric approaches have explored removing or modifying representations through subspace manipulation~\cite{gao2024ethos,uppaal2025model,wang2025model}, with complementary work investigating feature-space interventions using sparse autoencoders (SAEs).

UNLEARN~\cite{lizzo2025unlearn} introduced a subspace-based machine unlearning framework that identifies low-dimensional task representations and separates forget and retain components through subspace discrimination. Existing methods \cite{das2025} have largely relied on singular value decomposition (SVD) to recover these subspaces due to its optimal low-rank reconstruction properties. Rank-revealing QR (RRQR) factorizations, including column-pivoted and strong RRQR variants, provide a well-established alternative for identifying dominant low-dimensional subspaces and have been extensively studied in numerical linear algebra~\cite{hong1992rank,chandrasekaran1994rank,gu1996}, but have received comparatively little attention in machine unlearning.

\subsection{Cross-Lingual Machine Unlearning}

Cross-lingual unlearning remains relatively underexplored. Existing work has studied multilingual forgetting, misinformation removal, and transferability across languages~\cite{choi2024crosslingualunlearningselectiveknowledge,lu2025learnunlearnaddressingmisinformation,farashah2026multilingualamnesiatransferabilityunlearning}. More recent methods, including LING-TEA \cite{choi-etal-2024-cross}, explicitly leverage shared multilingual representations to improve cross-lingual forgetting. These observations motivate subspace-based approaches that directly manipulate shared multilingual parameter spaces, providing a natural mechanism for forgetting language-agnostic knowledge.

\subsection{Speech Foundation Models}

Self-supervised speech foundation models, including wav2vec~2.0~\cite{NEURIPS2020_92d1e1eb}, HuBERT~\cite{9585401}, and recent multimodal architectures such as Qwen3~\cite{xu2025qwen3omnitechnicalreport}, learn rich acoustic, linguistic, and speaker representations. Benchmarks such as VoxCeleb~\cite{Nagrani_2017,Chung_2018} provide controlled evaluation of both sample- and speaker-level forgetting, enabling assessment of whether machine unlearning methods generalize beyond text.

\section{Methodology}
\textsc{QR-Erase} is a subspace-based machine unlearning method that removes targeted knowledge by identifying and suppressing low-dimensional directions in model parameter space. Given a forget set and a retain set, \textsc{QR-Erase} first constructs task matrices that capture parameter updates associated with the corresponding data partitions. A column-pivoted QR (CPQR) decomposition identifies orthogonal bases for the resulting task subspaces, after which components shared with the retain set are removed through subspace discrimination. The remaining task-specific directions are then subtracted directly from the model parameters, yielding an efficient, interpretable machine unlearning procedure.

\subsection{\textsc{QR-Erase}}

Let $W^\ell$ denote the parameters of layer $\ell$. Following task-vector construction, we obtain layer-wise task matrices $T_f^\ell$ and $T_r^\ell$ corresponding to the forget and retain sets, respectively.

To identify the dominant directions associated with each set, we compute column-pivoted QR factorizations
\begin{equation}
T_f^\ell \Pi_f^\ell
=
Q_f^\ell R_f^\ell,
\label{eq:qr_forget}
\end{equation}
and
\begin{equation}
T_r^\ell \Pi_r^\ell
=
Q_r^\ell R_r^\ell,
\label{eq:qr_retain}
\end{equation}
where $Q_f^\ell$ and $Q_r^\ell$ contain orthonormal bases for the corresponding subspaces, $R_f^\ell$ and $R_r^\ell$ are upper-triangular matrices, and $\Pi_f^\ell$ and $\Pi_r^\ell$ are permutation matrices generated by column pivoting. We compute these factorizations using the standard CPQR implementation in \texttt{scipy.linalg.qr} with column pivoting enabled, which interfaces with LAPACK's \texttt{xGEQP3} routine. Our implementation therefore uses standard CPQR rather than the strong RRQR algorithm of Gu and Eisenstat~\cite{gu1996}.

Since the columns of $Q_r^\ell$ form an orthonormal basis for the retain subspace, the corresponding projection matrix is
\begin{equation}
P_r^\ell
=
Q_r^\ell (Q_r^\ell)^T.
\label{eq:retain_projection}
\end{equation}

To preserve retained knowledge, \textsc{QR-Erase} removes from the forget subspace all components that lie within the retain subspace:
\begin{equation}
\widetilde{Q}_f^\ell
=
(I-P_r^\ell)Q_f^\ell.
\label{eq:discrimination}
\end{equation}
This operation isolates directions that are unique to the forget set while preserving directions shared between the forget and retain sets.

The discriminated task matrix is reconstructed as
\begin{equation}
\widetilde{T}_f^\ell
=
\widetilde{Q}_f^\ell
R_f^\ell
(\Pi_f^\ell)^T.
\label{eq:reconstruction}
\end{equation}

Model parameters are then updated according to
\begin{equation}
(W^\ell)'
=
W^\ell
-
 \widetilde{T}_f^\ell
\label{eq:forget_update}
\end{equation}
Because column pivoting orders basis vectors according to their contribution to the underlying subspace, lower-rank approximations can be formed by retaining only the leading columns of $Q_f^\ell$ and the corresponding rows of $R_f^\ell$. This property enables efficient forgetting using compact low-dimensional representations.

\subsection{Theoretical Motivation}
\label{sec:theory}

The objective of \textsc{QR-Erase} is to identify and manipulate task-specific subspaces. Consequently, the quality of the decomposition should be evaluated primarily in terms of subspace recovery rather than matrix reconstruction error. Let
\begin{equation}
T_f^\ell \Pi_f^\ell
=
\begin{bmatrix}
Q_{f,k}^\ell &
Q_{f,\perp}^\ell
\end{bmatrix}
\begin{bmatrix}
R_{11} & R_{12} \\
0 & R_{22}
\end{bmatrix},
\label{eq:rrqr}
\end{equation}
denote the column-pivoted QR (CPQR) factorization of the forget task matrix, partitioned at rank $k$. Let
\begin{equation}
\mathcal{F}
=
\mathcal{R}(T_f^\ell)
\end{equation}
denote the original forget subspace and
\begin{equation}
\widehat{\mathcal{F}}=\mathcal{R}(Q_{f,k}^\ell)
\end{equation}
denote its recovered rank-$k$ approximation. A natural measure of subspace quality is the largest principal angle $\Theta(\mathcal{F},\widehat{\mathcal{F}})$ between the original and recovered subspaces. Classical results on rank-revealing QR factorizations relate this quantity to the singular value spectrum of the task matrix \cite{gu1996}. In particular,
\begin{equation}
\sin \Theta(\mathcal{F},\widehat{\mathcal{F}})\le \sigma_{k+1}\|R_{11}^{-1}\|_2,
\label{eq:principal_angle_bound}
\end{equation}
where $\sigma_{k+1}$ is the $(k+1)$-st singular value of $T_f^\ell$. The principal-angle distance is equivalently expressed as
\begin{equation}
\sin \Theta(\mathcal{F},\widehat{\mathcal{F}})=\|P_{\mathcal{F}}-P_{\widehat{\mathcal{F}}}\|_2,
\label{eq:projector_distance}
\end{equation}
where $P_{\mathcal{F}}$ and $P_{\widehat{\mathcal{F}}}$ denote the orthogonal projection operators onto the corresponding subspaces. Assuming the task matrix exhibits a spectral gap,
\begin{equation}
\sigma_{k+1}\ll\sigma_k,
\label{eq:spectral_gap}
\end{equation}
classical RRQR results indicate that the leading triangular factor preserves the dominant singular directions, yielding the approximation
\begin{equation}
\sin\Theta(\mathcal{F},\widehat{\mathcal{F}})=\mathcal{O}\!\left(\frac{\sigma_{k+1}}{\sigma_k}\right).
\label{eq:spectral_ratio}
\end{equation}
This relationship suggests that, when a sufficient spectral gap exists, CPQR recovers a subspace whose projection operator closely approximates that of the original forget subspace. Consequently, the update in Equation~\ref{eq:forget_update} is expected to suppress the dominant task-specific directions while preserving information shared with the retain subspace.

\subsubsection{Implications for Forgetting}
The preceding analysis provides theoretical motivation for using CPQR as the subspace recovery mechanism within \textsc{QR-Erase}. While SVD yields the optimal low-rank approximation in terms of reconstruction error, effective machine unlearning depends on identifying the dominant forget subspace rather than exactly reconstructing the task matrix. The extent to which this approximation is sufficient for practical machine unlearning is evaluated empirically in later sections.

\subsection{Layer-Localized \textsc{QR-Erase}}

Prior subspace-based machine unlearning methods apply forgetting uniformly across all modified layers~\cite{lizzo2025unlearn}, implicitly assuming that task-specific information is evenly distributed throughout the network. However, our empirical analysis suggests that task information is concentrated within a relatively small subset of layers. To exploit this structure, we introduce Layer-Localized \textsc{QR-Erase}, which restricts forgetting operations to layers contributing above-average task energy.

For each layer, we define the normalized task energy
\begin{equation}
E_\ell
=
\frac{\|T_f^\ell\|_F}
{\sum_{j\in\mathcal{L}}\|T_f^j\|_F},
\label{eq:energy}
\end{equation}
where $\mathcal{L}$ denotes the set of candidate layers. Since the normalized energies sum to one, the average contribution of a layer is simply $1/|\mathcal{L}|$. We therefore select layers whose task energy exceeds this average:
\begin{equation}
\mathcal{L}_{\mathrm{forget}}=\left\{\ell \in \mathcal{L}:E_\ell\ge\frac{1}{|\mathcal{L}|}\right\}.
\label{eq:layer_selection}
\end{equation}

The update in Equation~\ref{eq:forget_update} is then applied only to the selected layers,
\begin{equation}
(W^\ell)'
=
W^\ell
-
\widetilde{T}_f^\ell,
\qquad
\ell \in \mathcal{L}_{\mathrm{forget}},
\label{eq:localized_forget_update}
\end{equation}
while all remaining layers are left unchanged.

\section{Experimental Setup}
\label{sec:experimental_setup}

To evaluate the generality of \textsc{QR-Erase}, we consider three complementary machine unlearning settings: monolingual text unlearning, cross-lingual text unlearning, and speech unlearning. Unless otherwise noted, all experiments follow the evaluation protocols and hyperparameters of their respective benchmark papers to enable direct comparison with prior work.

\subsection{Text Unlearning}

We evaluate \textsc{QR-Erase} on two standard text-unlearning benchmarks: the original \textsc{UNLEARN} benchmark~\cite{lizzo2025unlearn}, consisting of arithmetic and reasoning tasks drawn from HELM~\cite{liang2023holistic}, and the TOFU benchmark~\cite{maini2024tofutaskfictitiousunlearning}, which measures selective forgetting of synthetic factual knowledge.

Experiments on the original \textsc{UNLEARN} benchmark use Llama-3-70B with LoRA-based task-matrix construction following~\cite{lizzo2025unlearn}. We compare against the optimization-based baselines reported in the original benchmark: Gradient Ascent~\cite{jang2023knowledge}, KGA~\cite{wang2023kga}, and KU~\cite{jang2022knowledge}, together with the SVD-based \textsc{UNLEARN}. For TOFU, we use the standard 10\% forget setting and original evaluation protocol, comparing against Gradient Ascent~\cite{jang2023knowledge}, Gradient Difference~\cite{liu2022continual}, Negative Preference Optimization~\cite{zhang2024npo}, FLAT~\cite{wang2024flat}, and the SVD-based \textsc{UNLEARN}~\cite{lizzo2025unlearn}. Layer-Localized variants of \textsc{UNLEARN} and \textsc{QR-Erase} are evaluated on both benchmarks. Performance is reported using the benchmark-specific metrics from the original works.

\begin{table}
    \centering
    \caption{
    Task-level text unlearning results when removing \textsc{Arithmetic}. Higher values indicate better retained-task performance, while lower \textsc{Arithmetic} performance indicates stronger forgetting. Bold and underlined values denote the best and second-best results.
    }
    \label{tab:text_task_results}
    \begin{tabular}{lccc}
        \hline
        \textbf{Method}
        & \textbf{MMLU} $\uparrow$
        & \textbf{MATH} $\uparrow$
        & \textbf{Arithmetic} $\downarrow$ \\
        \hline
        Base Model
        & 0.796
        & 0.505
        & 0.995 \\
        \hline
        Gradient Ascent
        & \underline{0.788}
        & 0.225
        & 0.084 \\
        KGA
        & 0.773
        & 0.110
        & \textbf{0.017} \\
        KU
        & 0.774
        & 0.191
        & \underline{0.063} \\
        \textsc{UNLEARN}
        & 0.777
        & \underline{0.482}
        & 0.828 \\
        \textsc{UNLEARN (LL)}
        & 0.784
        & \textbf{0.484}
        & 0.201 \\
        \textsc{QR-Erase}
        & 0.780
        & 0.475
        & 0.813 \\
        \textsc{QR-Erase} (LL)
        & \textbf{0.791}
        & 0.481
        & 0.203 \\
        \hline
    \end{tabular}
\end{table}

\begin{table}[t]
    \centering
    \caption{
    Results on the 10\% TOFU forget setting. Higher forget-quality $p$-values indicate stronger agreement with the retain-only reference model, while higher relative utility indicates better preservation of retained capabilities. Checkmarks indicate successful forgetting ($p>0.1$). Bold and underlined values denote the best and second-best results.
    }
    \label{tab:tofu_results}
    \begin{tabular}{lcc}
        \hline
        \textbf{Method}
        & \textbf{Model Utility} $\uparrow$
        & \textbf{Forget Quality} $\uparrow$ \\
        \hline
        Base Model
        & 0.6227
        & --- \\
        \hline
        Gradient Ascent
        & 0
        & $1.433{\times}10^{-22}$ \\   
        Gradient Difference
        & 0
        & $2.828 {\times}10^{-25}$ \\
        NPO
        & 0.2297
        & 0.0291 \\
        FLAT
        & \underline{0.6104}
        & 0.0841 \\
        \textsc{UNLEARN}
        & 0.6047
        & 0.257$^{\checkmark}$ \\
        \textsc{UNLEARN (LL)}
        & 0.6103
        & \textbf{0.281}$^{\checkmark}$ \\
        \textsc{QR-Erase}
        & 0.5938
        & 0.243$^{\checkmark}$ \\
        \textsc{QR-Erase} (LL)
        & \textbf{0.6120}
        & \underline{0.279}$^{\checkmark}$ \\
        \hline
    \end{tabular}
\end{table}

\subsection{Cross-Lingual Unlearning}

We further evaluate \textsc{QR-Erase} on the cTOFU benchmark~\cite{lizzo2026crosslingual}, a multilingual extension of TOFU. We adopt the released dataset and experimental protocol from~\cite{lizzo2026crosslingual} without modification.

Experiments use the standard 10\% forget setting and the multilingual model families evaluated in the original study. We compare \textsc{QR-Erase} and Layer-Localized \textsc{QR-Erase} against the same optimization- and subspace-based baselines. Performance is measured using the TOFU Truth Ratio, KS forgetting criterion, and relative model utility.

\begin{table*}[t]
\centering
\small
\setlength{\tabcolsep}{5pt}
\caption{
Cross-lingual unlearning results on cTOFU using Llama~3. Unlearning is performed in English and evaluated across four languages. Higher utility and forget quality indicate better performance. Checkmarks indicate successful forgetting ($p>0.1$). Bold and underlined values denote the best and second-best results. Comprehensive results are provided in the Appendix.
}
\label{tab:crosslingual_results}
\begin{tabular}{lcccccccc}
\toprule
\multirow{2}{*}{\textbf{Method}}
& \multicolumn{2}{c}{\textbf{English}}
& \multicolumn{2}{c}{\textbf{Spanish}}
& \multicolumn{2}{c}{\textbf{Italian}}
& \multicolumn{2}{c}{\textbf{Hindi}} \\
\cmidrule(lr){2-3}\cmidrule(lr){4-5}\cmidrule(lr){6-7}\cmidrule(lr){8-9}
& Utility $\uparrow$ & Forget $\uparrow$
& Utility $\uparrow$ & Forget $\uparrow$
& Utility $\uparrow$ & Forget $\uparrow$
& Utility $\uparrow$ & Forget $\uparrow$ \\
\midrule
Baseline
& 0.6227 & ---
& 0.5919 & ---
& 0.6102 & ---
& 0.5293 & --- \\

\midrule
NPO
& 0.2297 & 0.0291
& 0.0941 & $5.63{\times}10^{-14}$
& 0.0554 & $3.118{\times}10^{-16}$
& 0.0473 & $5.417{\times}10^{-21}$ \\

FLAT
& \underline{0.6104} & 0.0841
& 0.4174 & $1.73{\times}10^{-12}$
& 0.4872 & $6.28{\times}10^{-9}$
& 0.3399 & $8.92{\times}10^{-22}$ \\

UNLEARN
& 0.6047 & 0.257$^{\checkmark}$
& 0.5105 & 0.197$^{\checkmark}$
& 0.5173 & 0.249$^{\checkmark}$
& 0.4103 & 0.174$^{\checkmark}$ \\

UNLEARN (LL)
& 0.6103 & \textbf{0.281}$^{\checkmark}$
& \textbf{0.5303} & \textbf{0.218}$^{\checkmark}$
& \textbf{0.5736} & \underline{0.254}$^{\checkmark}$
&\textbf{ 0.4437} & \textbf{0.187}$^{\checkmark}$ \\

\textsc{QR-Erase}
& 0.5938
& 0.243$^{\checkmark}$
& 0.5006&0.203$^{\checkmark}$
&0.5016&0.235$^{\checkmark}$
&0.3852&0.153$^{\checkmark}$\\

\textsc{QR-Erase} (LL)
& \textbf{0.6120}
& \underline{0.279}$^{\checkmark}$
&\underline{0.5268}&\underline{0.214}$^{\checkmark}$
&\underline{0.5663}&\textbf{0.257}$^{\checkmark}$
&\underline{0.4391}&\underline{0.182}$^{\checkmark}$\\

\bottomrule
\end{tabular}
\end{table*}

\begin{figure}[t]
    \centering
    \includegraphics[
        width=0.70\linewidth,
        clip
    ]{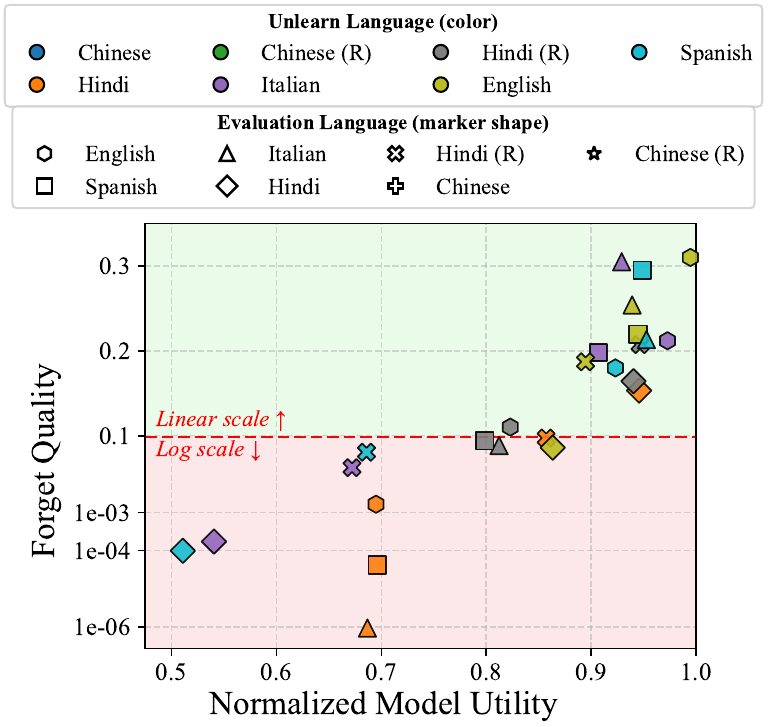}
    \caption{
    Cross-lingual forgetting and normalized model utility for \textsc{QR-Erase} on Llama~3 across language pairs. Marker color denotes the unlearning language, and marker shape denotes the evaluation language. Full results for Mixtral and Aya~23 are provided in the Appendix.
    }
    \label{fig:crosslingual_results}
\end{figure}

\subsection{Speech Unlearning}

Finally, we evaluate \textsc{QR-Erase} on speech foundation models using the VoxCeleb2 benchmark~\cite{Chung_2018}. Following~\cite{cheng25d_interspeech}, experiments are conducted using HuBERT~\cite{9585401} and Qwen3-Omni~\cite{xu2025qwen3omnitechnicalreport} after fine-tuning for speaker identification.

We consider two complementary tasks. In \emph{sample unlearning}, randomly selected utterances are removed from the training set, while in \emph{class unlearning}, all utterances associated with selected speaker identities are removed. Both settings use a 10\% forget partition.

We compare \textsc{QR-Erase} and Layer-Localized \textsc{QR-Erase} against Gradient Ascent~\cite{jang2023knowledge}, SCRUB~\cite{scrub2023}, Bad Teaching~\cite{badteaching2023}, and the SVD-based \textsc{UNLEARN}~\cite{lizzo2025unlearn}. Performance is evaluated using speaker identification top-5 accuracy on the forget and retain sets.

\section{Results}
\label{sec:results}

We evaluate \textsc{QR-Erase} across monolingual text, cross-lingual text, and speech unlearning, comparing against representative optimization-based baselines, the original SVD-based approach, and Layer-Localized \textsc{QR-Erase}.

\subsection{Text Unlearning}
\label{sec:text_results}

\subsubsection{Task-Level Knowledge Removal}

Table~\ref{tab:text_task_results} summarizes task-level unlearning results when removing the \textsc{Arithmetic} benchmark while preserving the closely related \textsc{MATH} benchmark. Optimization-based approaches successfully reduce arithmetic performance but introduce substantial collateral degradation, reducing \textsc{MATH} accuracy by more than 55\%.

In contrast, the subspace-based approaches preserve nearly all \textsc{MATH} performance while selectively removing arithmetic knowledge. \textsc{QR-Erase} closely matches the SVD-based \textsc{UNLEARN} baseline, demonstrating that accurate task-subspace recovery is sufficient for effective unlearning. Applying layer-localized forgetting further improves the forgetting--retention tradeoff for both decomposition methods, reducing arithmetic accuracy from 99.5\% to approximately 20\% while preserving nearly all \textsc{MATH} performance. These results indicate that identifying where task-specific knowledge is represented has a greater impact on selective forgetting than the particular subspace decomposition.

\subsubsection{TOFU Factual Unlearning}

Table~\ref{tab:tofu_results} reports results on the standard 10\% TOFU forget setting. Optimization-based methods exhibit a clear tradeoff between forget quality and model utility. While Gradient Ascent and Gradient Difference effectively remove the target knowledge, they do so at the cost of catastrophic utility degradation. Conversely, NPO and FLAT preserve substantially more utility but fail to satisfy the TOFU forgetting criterion.

The subspace-based approaches achieve a substantially stronger forgetting--utility tradeoff. \textsc{UNLEARN}, \textsc{QR-Erase}, and their layer-localized variants all satisfy the TOFU forgetting criterion while maintaining utility near the pretrained baseline. Replacing SVD with Pivoted QR results in only minor differences in forget quality and utility, further demonstrating that accurate subspace recovery is sufficient for effective machine unlearning. Applying layer-localized forgetting consistently improves the forgetting--utility tradeoff for both decomposition methods, with Layer-Localized \textsc{QR-Erase} achieving the highest model utility (0.6120) and Layer-Localized \textsc{UNLEARN} obtaining the highest forget quality ($p=0.281$). Together, these results suggest that factual knowledge is concentrated within a subset of model layers that can be selectively modified.

\begin{table*}[t]
    \centering
    \caption{
    Unlearning results on VoxCeleb2 for sample and class unlearning using HuBERT and Qwen3-Omni. $D_t$, $D_f$, and $D_r$ denote target-task, forget-task, and retain-task accuracy, respectively. Lower $D_f$ indicates better forgetting, while higher $D_t$ and $D_r$ indicate better preservation of retained capabilities. Bold and underlined values denote the best and second-best results.
    }
    \label{tab:speech_results}
    \begin{tabular}{lccc ccc ccc ccc}
        \hline
        \textbf{Method}
        & \multicolumn{3}{c}{\textbf{HuBERT Sample}}
        & \multicolumn{3}{c}{\textbf{HuBERT Class}}
        & \multicolumn{3}{c}{\textbf{Qwen3-Omni Sample}}
        & \multicolumn{3}{c}{\textbf{Qwen3-Omni Class}} \\
        \cmidrule(lr){2-4}
        \cmidrule(lr){5-7}
        \cmidrule(lr){8-10}
        \cmidrule(lr){11-13}
        & $D_t \uparrow$ & $D_f \downarrow$ & $D_r \uparrow$
        & $D_t \uparrow$ & $D_f \downarrow$ & $D_r \uparrow$
        & $D_t \uparrow$ & $D_f \downarrow$ & $D_r \uparrow$
        & $D_t \uparrow$ & $D_f \downarrow$ & $D_r \uparrow$ \\
        \hline
        Fine-Tuned Model
        & 84.1 & 84.4 & 83.7
        & 73.5 & 73.1 & 73.4
        & 94.1 & 94.4 & 93.7
        & 87.5 & 87.1 & 87.4 \\
        \hline
        Gradient Ascent
        & 14.1 & 13.4 & 17.4
        & 14.3 & 12.6 & 15.7
        & 15.9 & 17.3 & 21.7
        & 22.6 & 24.3 & 26.1 \\
        SCRUB
        & 48.2 & 51.1 & 50.2
        & 33.2 & 34.4 & 34.4
        & 54.8 & 58.3 & 52.1
        & 48.2 & 51.7 & 48.4 \\
        Bad Teaching
        & 53.7 & 51.6 & 55.9
        & 43.6 & 42.3 & 46.9
        & 59.3 & 62.4 & 59.3
        & 58.3 & 53.7 & 59.9 \\
        \textsc{UNLEARN}
        & 71.8 & 40.5 & 69.8
        & \underline{71.6} & 11.8 & \textbf{71.6}
        & 84.5 & 53.1 & \underline{86.3}
        & \underline{83.8} & 19.5 & 81.5 \\
        \textsc{UNLEARN (LL)}
        & \underline{73.4} & \textbf{15.3} & \underline{72.7}
        & \textbf{71.8} & \textbf{10.9} & \underline{71.2}
        & \underline{84.9} & \underline{16.5} & \textbf{87.1}
        & \textbf{84.2} & \underline{16.6} & \textbf{83.7} \\
        \textsc{QR-Erase}
        & \textbf{73.8} & 40.9 & \textbf{73.1}
        & 69.3 & 11.4 & 70.9
        & 83.7 & 52.5 & 84.6
        & 82.6 & 19.3 & 81.9 \\
        \textsc{QR-Erase} (LL)
        & 72.5 & \underline{16.7} & 69.6
        & 70.0 & \underline{11.3} & 69.4
        & \textbf{85.9} & \textbf{15.7} & \underline{86.3}
        & \underline{84.0} & \textbf{16.4} & \underline{83.3} \\
        \hline
    \end{tabular}
\end{table*}

\subsection{Cross-Lingual Unlearning}
\label{sec:crosslingual_results}

Table~\ref{tab:crosslingual_results} summarizes one-to-one cross-lingual unlearning on cTOFU. As in the monolingual setting, optimization-based methods exhibit a pronounced tradeoff between forget quality and model utility, with neither FLAT nor NPO consistently satisfying the TOFU forgetting criterion across the evaluated languages.

The subspace-based approaches achieve statistically significant forgetting while maintaining substantially higher utility across all languages. \textsc{QR-Erase} closely matches the SVD-based formulation, demonstrating that accurate subspace recovery generalizes to multilingual factual unlearning. Applying layer-localized forgetting further improves the forgetting--utility tradeoff for both decomposition methods, indicating that multilingual factual knowledge is concentrated within a subset of model layers rather than depending on the particular subspace decomposition.

\subsection{Speech Unlearning}
\label{sec:speech_results}

Table~\ref{tab:speech_results} summarizes sample and class unlearning results on VoxCeleb2. Optimization-based methods reduce forget-set accuracy but often substantially degrade target- and retain-set performance across both HuBERT and Qwen3-Omni.

In contrast, the subspace-based approaches preserve substantially more target- and retain-set performance while still reducing forget-set accuracy. Replacing SVD with Pivoted QR results in only minor differences across both speech models and unlearning settings, further demonstrating that accurate subspace recovery is sufficient for effective machine unlearning.

Applying layer-localized forgetting consistently improves the forgetting--retention tradeoff for both decomposition methods. On Qwen3-Omni sample unlearning, Layer-Localized \textsc{QR-Erase} reduces forget-set accuracy from 94.4\% to 15.7\% while preserving 85.9\% target-set accuracy and 86.3\% retain-set accuracy, whereas Layer-Localized \textsc{UNLEARN} achieves comparable performance with 16.5\% forget-set accuracy while attaining the highest retain-set performance (87.1\%). Similar improvements are observed for HuBERT, where both layer-localized methods substantially reduce forget-set accuracy relative to their non-localized counterparts while largely preserving target- and retain-set performance. These results indicate that speech representations, like textual knowledge, are concentrated within a subset of model layers that can be selectively modified.

\section{Analysis}
\label{sec:analysis}

The results across text, cross-lingual, and speech unlearning indicate that \textsc{QR-Erase} consistently identifies compact task-specific subspaces while preserving unrelated capabilities. We next analyze the approximation properties underlying this behavior and examine how task knowledge is distributed across model layers.

\subsection{Decomposition Runtime}
\label{sec:runtime}

To isolate decomposition cost, we compare the runtime of Pivoted QR and SVD on representative task-matrix dimensions spanning the language and speech models evaluated in this work. As shown in Table~\ref{tab:runtime}, Pivoted QR consistently reduces decomposition time by 65--68\% across matrix sizes ranging from HuBERT-scale ($768\times768$) to large LLM projection matrices ($8192\times8192$). For the largest matrix, decomposition time decreases from 2176.9\,s to 742.3\,s while maintaining comparable downstream unlearning performance. Since task-matrix construction is identical for both methods, these measurements isolate decomposition rather than end-to-end unlearning runtime. The results indicate that the exact low-rank optimality of SVD is unnecessary in practice, making Pivoted QR a computationally efficient alternative for subspace-based machine unlearning.

\begin{table}[t]
\centering
\caption{
Average CPU decomposition runtime for representative task-matrix dimensions.
Results are averaged over multiple runs after warm-up iterations.
Runtime reduction is computed relative to SVD.
}
\label{tab:runtime}
\begin{tabular}{lcccc}
\toprule
Matrix Size &
SVD (s) &
Pivoted QR (s) &
Reduction \\
\midrule
$768 \times 768$     & 1.84 & 0.60 & 67.3\% \\
$2048 \times 2048$   & 34.73 & 11.74 & 66.2\% \\
$4096 \times 4096$   & 277.85 & 92.41 & 66.7\% \\
$8192 \times 8192$   & 2176.90 & 742.34 & 65.9\% \\
\bottomrule
\end{tabular}
\end{table}

\subsection{Analysis of Low-Rank Approximation Quality}
\label{sec:approximation_analysis}

\begin{figure}[t]
    \centering
    \includegraphics[
        width=\linewidth,
        clip
    ]{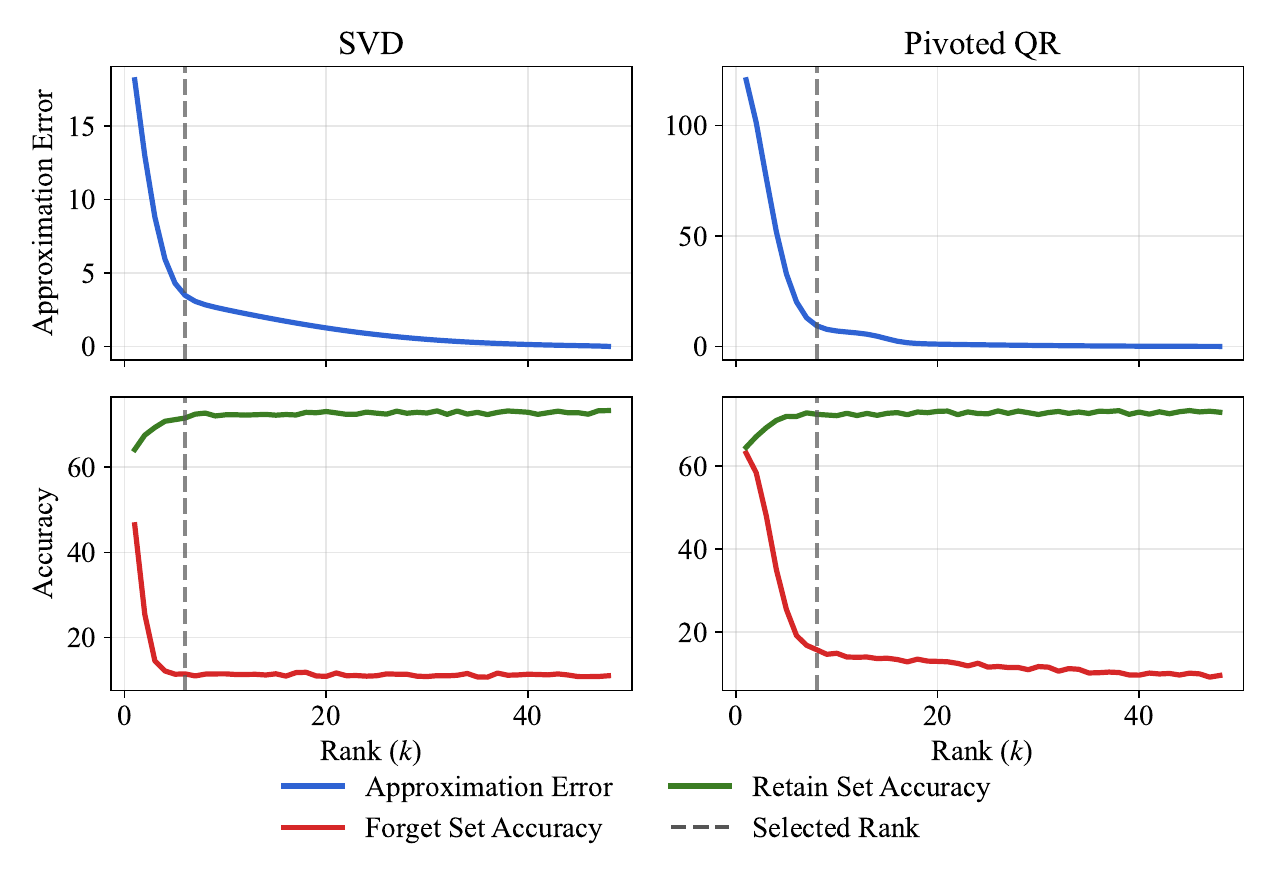}
    \caption{Representative rank-dependent approximation and downstream unlearning behavior for SVD and Pivoted QR (speech). Reconstruction error decreases with rank (top), while downstream forgetting saturates at low ranks (bottom). Dashed lines denote the selected ranks. Similar trends are observed for text and cross-lingual unlearning.}
    \label{fig:rank_error_accuracy}
\end{figure}

Although SVD provides the optimal rank-$k$ approximation, Figure~\ref{fig:rank_error_accuracy} shows that \textsc{QR-Erase} achieves nearly identical downstream unlearning performance. For each method, we select the smallest rank beyond which forget- and retain-set performance saturates, corresponding to $k=8$ for SVD and $k=10$ for Pivoted QR in the representative speech experiments. Although reconstruction error continues to decrease with rank, increasing the retained rank beyond these operating points changes forget- and retain-set performance by less than 1\%. Similar behavior is observed on TOFU and cTOFU, indicating that effective unlearning depends on recovering the dominant task subspace rather than the optimal low-rank reconstruction.

\subsection{Layer-wise Localization of Knowledge}
\label{sec:layer_localization_analysis}

\begin{figure}[t]
    \centering
    \includegraphics[
        width=.85\columnwidth,
        clip
    ]{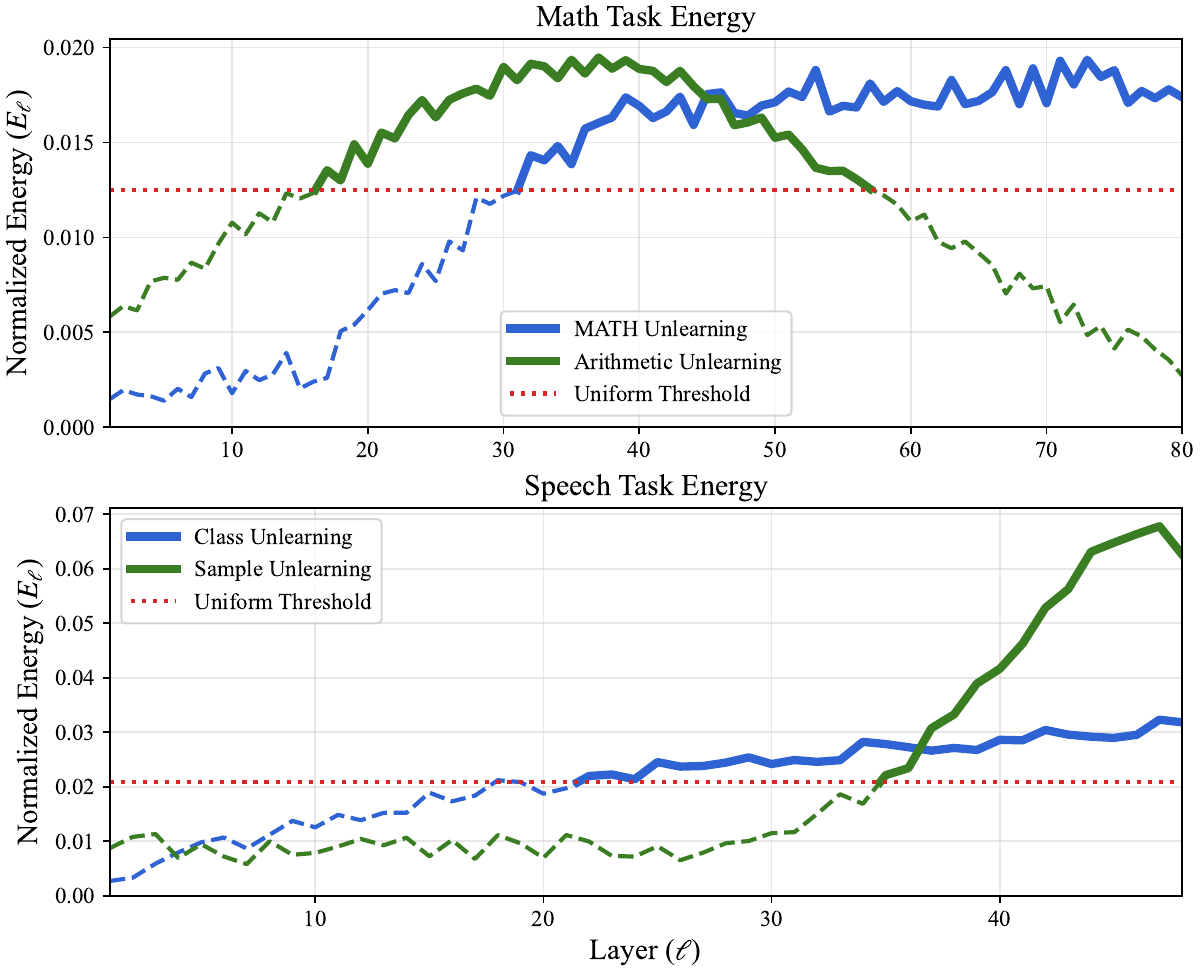}
    \caption{
    Layer-wise normalized task energy $E_{\ell}$ for representative unlearning tasks. The dashed horizontal line denotes the uniform baseline $1/|\mathcal{L}|$. Layers above this threshold are selected for layer-localized removal.
    }
    \label{fig:layer_energy}
\end{figure}

Figure~\ref{fig:layer_energy} shows the normalized layer-wise task energy for representative text and speech unlearning tasks. Across both domains, task-specific information is concentrated within a relatively small subset of layers, indicating that knowledge is spatially localized.

For speech, sample-level information is more localized than speaker identity, explaining why Layer-Localized \textsc{QR-Erase} achieves its largest improvements for sample unlearning (Table~\ref{tab:speech_results}). Similarly, the partially overlapping energy distributions of MATH and arithmetic enable more selective removal, reducing arithmetic performance from 81.3\% to 20.3\% while leaving MATH performance nearly unchanged (47.5\% versus 48.1\%).

Together, these results suggest that effective unlearning depends on both recovering the dominant task subspace and identifying where it is represented, enabling more targeted knowledge removal with less collateral degradation.

\section{Conclusion}

This work investigated two fundamental questions underlying subspace-based machine unlearning: whether optimal low-rank reconstruction is necessary for effective forgetting, and whether task-specific knowledge is uniformly distributed throughout a model. To study these questions, we introduced \textsc{QR-Erase}, a subspace-based machine unlearning framework that replaces singular value decomposition with computationally efficient Pivoted QR for task subspace recovery, together with a layer-localized forgetting strategy that restricts updates to layers containing above-average task energy.

Experimental results across task-level text unlearning, cross-lingual factual unlearning, and speech unlearning demonstrate that \textsc{QR-Erase} consistently matches or outperforms existing optimization- and subspace-based methods while preserving unrelated capabilities. Despite replacing singular value decomposition with Pivoted QR, \textsc{QR-Erase} remains within 5\% of the SVD-based approach across all evaluation settings while reducing decomposition runtime by over 60\%, demonstrating that accurate subspace recovery is sufficient for effective machine unlearning. In contrast, layer-localized forgetting consistently improves the forgetting--utility tradeoff for both QR- and SVD-based methods, indicating that identifying where task-specific knowledge is represented has a greater impact on unlearning performance than the particular subspace decomposition.

Beyond introducing a computationally efficient unlearning framework, our analysis provides evidence that task-specific knowledge exhibits both low-rank structure and layer-wise localization across multiple modalities. Together, these findings suggest that effective machine unlearning depends on both identifying the appropriate task subspace and corresponding layers within the model.

Future work will investigate broader forms of knowledge, including multimodal concepts, reasoning capabilities, demographic attributes, and safety-critical behaviors, as well as adaptive strategies for identifying task-relevant subspaces and localized regions of knowledge. We hope these results encourage further exploration of efficient geometric approaches to scalable and interpretable machine unlearning.

\bibliography{aaai2027}

\appendix
\section{Method-Specific Settings}
\label{appendix:unlearning_setup}

Each baseline method is implemented using the objective and configuration described in the corresponding original work. Unless otherwise noted, hyperparameters follow the recommended settings from the original publications.

\paragraph{Gradient Ascent:}
Learning rate of $5\times10^{-5}$, batch size equivalent to the size of the forget set, and $EL_n$ with $n=10$.

\paragraph{Knowledge Unlearning (KU):}
Implemented using the objective and hyperparameters from~\cite{jang2022knowledge}.

\paragraph{Knowledge Gap Alignment (KGA):}
Implemented using the objective and hyperparameters from~\cite{wang2023kga}.

\paragraph{KL-based Retain Objective:}
Learning rate of $10^{-5}$, batch size of 32, 5 training epochs, and $\alpha=8$.

\paragraph{Gradient Difference:}
Learning rate of $10^{-5}$, batch size of 32, and 5 training epochs.

\paragraph{Negative Preference Optimization (NPO):}
Learning rate of $10^{-5}$, batch size of 32, and 5 training epochs.

\paragraph{Direct Preference Optimization (DPO):}
Learning rate of $10^{-6}$, $\beta=0.1$, and batch size of 64.

\paragraph{FLAT:}
Learning rate of $10^{-5}$, batch size of 32, and 5 training epochs.

\paragraph{SCRUB:}
Implemented following the original SCRUB objective~\cite{scrub2023}. Unless otherwise noted, we use the optimization procedure described in the original work, consisting of alternating forget (max) and retain (min) optimization phases with Adam. Hyperparameters follow the recommendations of the original implementation.
\paragraph{Bad Teaching:}
Implemented following the original teacher--student framework of~\cite{badteaching2023}. We use the Adam optimizer with a single unlearning epoch and a randomly initialized incompetent teacher paired with the original fine-tuned model as the competent teacher. Unless otherwise noted, remaining hyperparameters follow the recommendations of the original implementation.

\paragraph{\textsc{UNLEARN}:}
Learning rate of $10^{-5}$, batch size of 32, 5 training epochs, and subspace rank $k=4$. Task vectors are constructed using LoRA fine-tuning as described in~\cite{lizzo2025unlearn}. Layer-Localized \textsc{UNLEARN} uses the same configuration, with updates restricted to the layers selected by Equation~\ref{eq:layer_selection}.

\paragraph{\textsc{QR-Erase}:}
Unless otherwise noted, \textsc{QR-Erase} follows the same training configuration as \textsc{UNLEARN}, including LoRA task-vector construction, learning rate of $10^{-5}$, batch size of 32, 5 training epochs, and subspace rank $k=4$. The task subspace is recovered using the column-pivoted QR (CPQR) implementation provided by \texttt{scipy.linalg.qr} with \texttt{pivoting=True}, which interfaces with LAPACK's \texttt{xGEQP3} routine. Layer-Localized \textsc{QR-Erase} applies the same procedure while restricting updates to the layers selected by Equation~\ref{eq:layer_selection}.

\subsection{Decomposition Runtime Benchmark}
\label{app:runtime}

Runtime measurements were performed on a CPU-only system with an
Intel\textsuperscript{\textregistered} Core\texttrademark~i7-4578U processor
(2 physical cores, 4 hardware threads) and 8\,GB of system memory.
Benchmarks were implemented in Python using NumPy and SciPy. Both SVD and
Pivoted QR used the corresponding routines from \texttt{scipy.linalg}
(\texttt{svd} and \texttt{qr} with \texttt{pivoting=True}), operating on
double-precision (\texttt{float64}) matrices stored in Fortran-contiguous
memory layout.

For each matrix size, a random matrix was generated from a standard normal
distribution using a fixed random seed. Each decomposition was preceded by
warm-up iterations to mitigate initialization overhead, followed by multiple
timed runs using \texttt{time.perf\_counter()}. Reported runtimes correspond
to the average wall-clock time across the timed runs. The benchmark isolates
only the decomposition stage and does not include task-matrix construction or
other components of the unlearning pipeline.

\begin{table}[t]
\centering
\small
\begin{tabular}{l l}
\toprule
\textbf{Model} & \textbf{HuggingFace Checkpoint} \\
\midrule
Llama 3 70B
& \texttt{meta-llama/Meta-Llama-3-70B} \\

Llama 3 8B
& \texttt{meta-llama/Meta-Llama-3-8B} \\

Mixtral 8$\times$7B
& \texttt{mistralai/Mixtral-8x7B-v0.1} \\

Aya 23 8B
& \texttt{CohereForAI/aya-23-8B} \\

HuBERT XLarge
& \texttt{facebook/hubert-xlarge-ll60k} \\

Qwen3-Omni
& \texttt{Qwen/Qwen3-Omni-30B-A3B-Instruct} \\
\bottomrule
\end{tabular}
\caption{Pretrained HuggingFace checkpoints used throughout the text, cross-lingual, and speech unlearning experiments.}
\label{tab:hf_checkpoints}
\end{table}

\section{Crosslingual Full Results}
\label{appendix:crosslingual_results}
To complement the representative results presented in the main paper, this section provides the complete cross-lingual evaluation. Table~\ref{tab:crosslingual_full} summarizes one-to-one cross-lingual unlearning across all evaluated multilingual foundation models and baseline methods. Table~\ref{tab:crosslingual_all_pairs} reports the complete language-pair results for Layer-Localized \textsc{QR-Erase}, including every evaluated unlearning and evaluation language combination.

\begin{table*}[t]
\centering
\small
\setlength{\tabcolsep}{5pt}
\begin{tabular}{|l|cc|cc|cc|cc|}
\hline
\textbf{Method}
& \multicolumn{2}{c|}{\textbf{en$\to$en}}
& \multicolumn{2}{c|}{\textbf{en$\to$es}}
& \multicolumn{2}{c|}{\textbf{es$\to$en}}
& \multicolumn{2}{c|}{\textbf{es$\to$es}} \\
& Utility $\uparrow$ & Forget $\uparrow$
& Utility $\uparrow$ & Forget $\uparrow$
& Utility $\uparrow$ & Forget $\uparrow$
& Utility $\uparrow$ & Forget $\uparrow$ \\
\hline

\multicolumn{9}{|l|}{\textbf{Llama 3}} \\ \hline
Base Model
& 0.6227 & $2.37{\times}10^{-15}$
& 0.5919 & $4.31{\times}10^{-13}$
& 0.6227 & $2.37{\times}10^{-15}$
& 0.5919 & $4.31{\times}10^{-13}$ \\
\hline
DPO
& 0.5357 & $1.06{\times}10^{-16}$
& 0.0506 & $1.42{\times}10^{-18}$
& 0.1025 & $3.58{\times}10^{-12}$
& 0.5216 & $2.84{\times}10^{-18}$ \\
Gradient Ascent
& 0.0000 & $1.35{\times}10^{-23}$
& 0.0000 & $7.74{\times}10^{-22}$
& 0.0000 & $2.98{\times}10^{-23}$
& 0.0000 & $3.25{\times}10^{-27}$ \\
KL Minimization
& 0.0000 & $4.07{\times}10^{-32}$
& 0.0478 & $2.63{\times}10^{-17}$
& 0.0000 & $2.91{\times}10^{-11}$
& 0.0467 & $7.84{\times}10^{-18}$ \\
Gradient Difference
& 0.5872 & $2.83{\times}10^{-25}$
& 0.0484 & $3.74{\times}10^{-22}$
& 0.1136 & $1.47{\times}10^{-18}$
& 0.5638 & $2.96{\times}10^{-19}$ \\
NPO
& 0.2297 & 0.0291
& 0.0941 & $5.63{\times}10^{-14}$
& 0.1173 & $9.46{\times}10^{-18}$
& 0.1639 & 0.0517 \\
FLAT
& \underline{0.6104} & 0.0841
& 0.4174 & $1.73{\times}10^{-12}$
& 0.3118 & $4.52{\times}10^{-15}$
& 0.5802 & 0.0712 \\
UNLEARN
& 0.6047 & 0.257$^{\checkmark}$
& 0.5105 & 0.197$^{\checkmark}$
& 0.5482 & \underline{0.182}$^{\checkmark}$
& \underline{0.5824} & 0.271$^{\checkmark}$ \\
UNLEARN (LL)
& 0.6103 & \textbf{0.281}$^{\checkmark}$
& \textbf{0.5303} & \textbf{0.218}$^{\checkmark}$
& \underline{0.5713} & \textbf{0.187}$^{\checkmark}$
& \textbf{0.5837} & \textbf{0.298}$^{\checkmark}$ \\
\textsc{QR-Erase}
& 0.5938 & 0.243$^{\checkmark}$
& 0.5006 & 0.203$^{\checkmark}$
& 0.5506 & 0.176$^{\checkmark}$
& 0.5585 & 0.267$^{\checkmark}$ \\
\textsc{QR-Erase} (LL)
& \textbf{0.6120} & \underline{0.279}$^{\checkmark}$
& \underline{0.5268} & \underline{0.214}$^{\checkmark}$
& \textbf{0.5748} & 0.180$^{\checkmark}$
& 0.5615 & \underline{0.295}$^{\checkmark}$ \\
\hline

\multicolumn{9}{|l|}{\textbf{Mixtral}} \\ \hline
Base Model
& 0.5912 & $3.48{\times}10^{-16}$
& 0.5393 & $6.91{\times}10^{-18}$
& 0.5912 & $3.48{\times}10^{-16}$
& 0.5393 & $6.91{\times}10^{-18}$ \\
\hline
DPO
& 0.5106 & $2.53{\times}10^{-14}$
& 0.0246 & $2.71{\times}10^{-17}$
& 0.0815 & $6.48{\times}10^{-14}$
& 0.4621 & $8.47{\times}10^{-17}$ \\
Gradient Ascent
& 0.0000 & $1.35{\times}10^{-19}$
& 0.0000 & $4.74{\times}10^{-28}$
& 0.0000 & $7.73{\times}10^{-23}$
& 0.0000 & $1.25{\times}10^{-30}$ \\
KL Minimization
& 0.0000 & $4.07{\times}10^{-19}$
& 0.0000 & $8.00{\times}10^{-29}$
& 0.0000 & $3.61{\times}10^{-17}$
& 0.0000 & $7.94{\times}10^{-21}$ \\
Gradient Difference
& 0.5483 & $5.83{\times}10^{-25}$
& 0.0587 & $8.92{\times}10^{-19}$
& 0.0439 & $9.74{\times}10^{-22}$
& 0.4914 & $3.64{\times}10^{-21}$ \\
NPO
& 0.1893 & 0.0587
& 0.0641 & $5.96{\times}10^{-19}$
& 0.0973 & $9.37{\times}10^{-22}$
& 0.2535 & 0.0219 \\
FLAT
& \textbf{0.5871} & 0.0692
& 0.2960 & $1.84{\times}10^{-22}$
& 0.2270 & $4.83{\times}10^{-22}$
& \textbf{0.5173} & 0.0174 \\
UNLEARN
& 0.5513 & 0.234$^{\checkmark}$
& 0.4495 & \underline{0.191}$^{\checkmark}$
& 0.4845 & \underline{0.163}$^{\checkmark}$
& 0.4984 & 0.210$^{\checkmark}$ \\
UNLEARN (LL)
& \underline{0.5812} & \textbf{0.257}$^{\checkmark}$
& \textbf{0.4679} & \textbf{0.208}$^{\checkmark}$
& \underline{0.5062} & \textbf{0.182}$^{\checkmark}$
& \underline{0.5004} & \textbf{0.271}$^{\checkmark}$ \\
\textsc{QR-Erase}
& 0.5576 & 0.237$^{\checkmark}$
& 0.4512 & 0.183$^{\checkmark}$
& 0.4797 & 0.151$^{\checkmark}$
& 0.4933 & 0.251$^{\checkmark}$ \\
\textsc{QR-Erase} (LL)
& 0.5794 & \textbf{0.241}$^{\checkmark}$
& \underline{0.4617} & \underline{0.191}$^{\checkmark}$
& \textbf{0.5082} & 0.159$^{\checkmark}$
& 0.4962 & \underline{0.260}$^{\checkmark}$ \\

\hline

\multicolumn{9}{|l|}{\textbf{Aya 23}} \\ \hline
Base Model
& 0.5503 & $6.47{\times}10^{-24}$
& 0.5193 & $2.63{\times}10^{-21}$
& 0.5503 & $6.47{\times}10^{-24}$
& 0.5193 & $2.63{\times}10^{-21}$ \\
\hline
DPO
& 0.4182 & $4.39{\times}10^{-21}$
& 0.0176 & $4.27{\times}10^{-22}$
& 0.0438 & $2.41{\times}10^{-19}$
& 0.3921 & $6.28{\times}10^{-15}$ \\
Gradient Ascent
& 0.0000 & $6.28{\times}10^{-21}$
& 0.0000 & $7.12{\times}10^{-26}$
& 0.0000 & $3.30{\times}10^{-27}$
& 0.0000 & $4.67{\times}10^{-33}$ \\
KL Minimization
& 0.0000 & $2.38{\times}10^{-21}$
& 0.0000 & $8.30{\times}10^{-23}$
& 0.0000 & $7.20{\times}10^{-26}$
& 0.0000 & $4.70{\times}10^{-21}$ \\
Gradient Difference
& 0.4978 & $6.19{\times}10^{-29}$
& 0.0589 & $8.94{\times}10^{-19}$
& 0.0439 & $9.36{\times}10^{-22}$
& 0.4914 & $3.58{\times}10^{-21}$ \\
NPO
& 0.1213 & 0.0468
& 0.0487 & $7.31{\times}10^{-21}$
& 0.0692 & $1.27{\times}10^{-19}$
& 0.2316 & 0.0382 \\
FLAT
& \textbf{0.5264} & 0.0765
& 0.2340 & $4.86{\times}10^{-28}$
& 0.1930 & $2.83{\times}10^{-23}$
& \textbf{0.4920} & 0.0234 \\
UNLEARN
& 0.5047 & 0.315$^{\checkmark}$
& 0.4210 & 0.248$^{\checkmark}$
& 0.4073 & \underline{0.192}$^{\checkmark}$
& 0.4621 & 0.272$^{\checkmark}$ \\
UNLEARN (LL)
& \underline{0.5164} & \textbf{0.331}$^{\checkmark}$
& \underline{0.4313} & \underline{0.269}$^{\checkmark}$
& \textbf{0.4316} & \textbf{0.193}$^{\checkmark}$
& \underline{0.4904} & \underline{0.279}$^{\checkmark}$ \\
\textsc{QR-Erase}
& 0.4957 & 0.276$^{\checkmark}$
& 0.4239 & 0.251$^{\checkmark}$
& 0.3993 & 0.183$^{\checkmark}$
& 0.4625 & 0.271$^{\checkmark}$ \\
\textsc{QR-Erase} (LL)
& 0.5115 & \underline{0.325}$^{\checkmark}$
& \textbf{0.4337} & \textbf{0.278}$^{\checkmark}$
& \underline{0.4299} & 0.189$^{\checkmark}$
& 0.4896 & \textbf{0.289}$^{\checkmark}$ \\

\hline
\end{tabular}
\caption{Cross-lingual factual unlearning results across three multilingual foundation models. Each column reports relative model utility (higher is better) and forget quality (higher is better) for the four combinations of unlearning and evaluation language. Baselines include Gradient Ascent~\cite{jang2023knowledge}, Gradient Difference~\cite{liu2022continual}, DPO~\cite{rafailov2024directpreferenceoptimizationlanguage}, KL Minimization~\cite{wang2025balancing}, NPO~\cite{zhang2024npo}, FLAT~\cite{wang2024flat}, and the subspace-based methods. Checkmarks indicate successful forgetting ($p>0.1$). Bold and underlined values denote the best and second-best results, respectively.}
\label{tab:crosslingual_full}
\end{table*}

\begin{table*}[t]
\centering
\scriptsize
\setlength{\tabcolsep}{4pt}
\begin{tabular}{ll rr rr rr}
\toprule
& & \multicolumn{2}{c}{\textbf{Llama 3}}
  & \multicolumn{2}{c}{\textbf{Mixtral}}
  & \multicolumn{2}{c}{\textbf{Aya 23}} \\
\cmidrule(lr){3-4}\cmidrule(lr){5-6}\cmidrule(lr){7-8}
Unlearn & Eval & Util. & FQ & Util. & FQ & Util. & FQ \\
\midrule

\multirow{7}{*}{Base Model}
& Chinese     & --- & --- & 0.05040 & $7.67{\times}10^{-17}$ & 0.2688 & $4.62{\times}10^{-19}$ \\
& Chinese (R) & --- & --- & 0.1693 & $7.88{\times}10^{-15}$ & 0.2983 & $2.82{\times}10^{-23}$ \\
& English     & 0.6227 & $2.00{\times}10^{-15}$ & 0.5912 & $2.00{\times}10^{-15}$ & 0.5503 & $6.40{\times}10^{-24}$ \\
& Hindi       & 0.4104 & $7.67{\times}10^{-17}$ & --- & --- & 0.3929 & $5.73{\times}10^{-21}$ \\
& Hindi (R)   & 0.5293 & $7.88{\times}10^{-15}$ & --- & --- & 0.4623 & $3.51{\times}10^{-17}$ \\
& Italian     & 0.6102 & $8.84{\times}10^{-17}$ & 0.5669 & $8.84{\times}10^{-17}$ & 0.5117 & $3.21{\times}10^{-19}$ \\
& Spanish     & 0.5919 & $4.31{\times}10^{-13}$ & 0.5393 & $4.31{\times}10^{-13}$ & 0.5193 & $2.60{\times}10^{-21}$ \\

\midrule
\multirow{7}{*}{English}
& English     & 0.6120 & 0.279 & 0.5794 & 0.241 & 0.5115 & 0.325 \\
& Spanish     & 0.5268 & 0.214 & 0.4617 & 0.191 & 0.4377 & 0.278 \\
& Italian     & 0.5663 & 0.257 & 0.4744 & 0.212 & 0.4607 & 0.278 \\
& Hindi       & 0.3543 & 0.0498 & --- & --- & 0.2696 & 0.0416 \\
& Hindi (R)   & 0.4391 & 0.182 & --- & --- & 0.3958 & 0.165 \\
& Chinese     & --- & --- & 0.04470 & $1.12{\times}10^{-11}$ & 0.1648 & 0.0469 \\
& Chinese (R) & --- & --- & 0.07990 & $5.89{\times}10^{-4}$ & 0.2669 & 0.173 \\

\midrule
\multirow{7}{*}{Spanish}
& English     & 0.5748 & 0.180 & 0.5082 & 0.159 & 0.4299 & 0.189 \\
& Spanish     & 0.5615 & 0.295 & 0.4962 & 0.260 & 0.4896 & 0.289 \\
& Italian     & 0.5813 & 0.214 & 0.4840 & 0.0430 & 0.4414 & 0.194 \\
& Hindi       & 0.2096 & $9.80{\times}10^{-5}$ & --- & --- & 0.1732 & 0.00140 \\
& Hindi (R)   & 0.3630 & 0.0377 & --- & --- & 0.4332 & 0.0712 \\
& Chinese     & --- & --- & 0.007200 & $9.87{\times}10^{-19}$ & 0.08350 & $6.74{\times}10^{-5}$ \\
& Chinese (R) & --- & --- & 0.004200 & $1.02{\times}10^{-10}$ & 0.2140 & 0.0826 \\

\midrule
\multirow{7}{*}{Italian}
& English     & 0.6057 & 0.212 & 0.5598 & 0.164 & 0.5309 & 0.237 \\
& Spanish     & 0.5368 & 0.198 & 0.5158 & 0.183 & 0.4753 & 0.195 \\
& Italian     & 0.5669 & 0.305 & 0.5228 & 0.260 & 0.4912 & 0.318 \\
& Hindi       & 0.2218 & $1.71{\times}10^{-4}$ & --- & --- & 0.1646 & 0.00396 \\
& Hindi (R)   & 0.3557 & 0.0149 & --- & --- & 0.3598 & 0.0506 \\
& Chinese     & --- & --- & 0.003500 & $1.29{\times}10^{-22}$ & 0.1334 & $8.31{\times}10^{-4}$ \\
& Chinese (R) & --- & --- & 0.006800 & $5.70{\times}10^{-12}$ & 0.1683 & 0.0648 \\

\midrule
\multirow{5}{*}{Hindi}
& English     & 0.4327 & 0.00163 & --- & --- & 0.4117 & 0.00845 \\
& Spanish     & 0.4119 & $4.15{\times}10^{-5}$ & --- & --- & 0.3188 & $6.72{\times}10^{-6}$ \\
& Italian     & 0.4190 & $9.23{\times}10^{-7}$ & --- & --- & 0.3076 & $1.87{\times}10^{-7}$ \\
& Hindi       & 0.3881 & 0.153 & --- & --- & 0.3361 & 0.200 \\
& Hindi (R)   & 0.4536 & 0.0891 & --- & --- & 0.4052 & 0.166 \\

\midrule
\multirow{5}{*}{Hindi (R)}
& English     & 0.5124 & 0.111 & --- & --- & 0.4716 & 0.141 \\
& Spanish     & 0.4726 & 0.0748 & --- & --- & 0.4190 & 0.0708 \\
& Italian     & 0.4956 & 0.0563 & --- & --- & 0.3807 & 0.0363 \\
& Hindi       & 0.3859 & 0.165 & --- & --- & 0.3358 & 0.185 \\
& Hindi (R)   & 0.5010 & 0.208 & --- & --- & 0.4432 & 0.225 \\

\midrule
\multirow{5}{*}{Chinese}
& English     & --- & --- & 0.05040 & $5.25{\times}10^{-9}$ & 0.3293 & $5.93{\times}10^{-4}$ \\
& Spanish     & --- & --- & 0.01460 & $1.29{\times}10^{-11}$ & 0.2726 & $1.07{\times}10^{-8}$ \\
& Italian     & --- & --- & 0.03240 & $2.12{\times}10^{-13}$ & 0.2379 & $9.73{\times}10^{-14}$ \\
& Chinese     & --- & --- & 0.01820 & $1.37{\times}10^{-5}$ & 0.2189 & 0.128 \\
& Chinese (R) & --- & --- & 0.08990 & $5.83{\times}10^{-5}$ & 0.2317 & 0.0907 \\

\midrule
\multirow{5}{*}{Chinese (R)}
& English     & --- & --- & 0.3739 & 0.00769 & 0.3550 & 0.0921 \\
& Spanish     & --- & --- & 0.2877 & 0.00384 & 0.2838 & 0.0645 \\
& Italian     & --- & --- & 0.2476 & 0.00135 & 0.2639 & 0.0436 \\
& Chinese     & --- & --- & 0.03500 & $1.01{\times}10^{-6}$ & 0.2220 & 0.119 \\
& Chinese (R) & --- & --- & 0.1324 & 0.0408 & 0.2729 & 0.189 \\

\bottomrule
\end{tabular}
\caption{Complete cross-lingual factual unlearning results for Layer-Localized \textsc{QR-Erase}. Rows correspond to the unlearning language and columns correspond to the evaluation language for \textbf{Llama 3}, \textbf{Mixtral}, and \textbf{Aya 23}. Each entry reports relative model utility (higher is better) and forget quality (higher is better). Romanized variants of Hindi and Chinese are denoted by ``(R).''}
\label{tab:crosslingual_all_pairs}
\end{table*}

\end{document}